\documentclass[journal,onecolumn,web]{IEEEtran}
\usepackage{cite}
\usepackage{amsmath, amssymb, amsfonts}
\usepackage{algorithmic}
\usepackage{graphicx}
\usepackage{textcomp}

\usepackage[hidelinks]{hyperref}
\usepackage{booktabs}
\usepackage{multirow}
\usepackage[table,xcdraw]{xcolor}
\usepackage[normalem]{ulem}
\useunder{\uline}{\ul}{}
\def\BibTeX{{\rm B\kern-.05em{\sc i\kern-.025em b}\kern-.08em
    T\kern-.1667em\lower.7ex\hbox{E}\kern-.125emX}}

\begin{document}
\title{Self-Supervised Multi-Modality Learning for Multi-Label Skin Lesion Classification}
\author{Hao Wang, \IEEEmembership{Student Member, IEEE}, Euijoon Ahn, \IEEEmembership{Member, IEEE}, Lei Bi, \IEEEmembership{Member, IEEE}, and Jinman Kim, \IEEEmembership{Member, IEEE}
\thanks{This work was supported in part by Australian Research Council (ARC) grants (DP200103748).}
\thanks{Hao Wang and Jinman Kim are with the School of Computer Science, University of Sydney, NSW, Australia (corresponding e-mail: jinman.kim@sydney.edu.au)}
\thanks{Euijoon Ahn is with the College of Science and Engineering, James Cook University, QLD, Australia.}
\thanks{Hao Wang and Lei Bi are with the Institute of Translational Medicine, National Center for Translational Medicine, Shanghai Jiao Tong University, Shanghai, China. (corresponding e-mail: lei.bi@sjtu.edu.cn).}
}

\maketitle

\begin{abstract}
The clinical diagnosis of skin lesion involves the analysis of dermoscopic and clinical modalities. Dermoscopic images provide a detailed view of the surface structures whereas clinical images offer a complementary macroscopic information. The visual diagnosis of melanoma is also based on seven-point checklist which involves identifying different visual attributes. Recently, supervised learning approaches such as convolutional neural networks (CNNs) have shown great performances using both dermoscopic and clinical modalities (Multi-modality). The seven different visual attributes in the checklist are also used to further improve the the diagnosis. The performances of these approaches, however, are still reliant on the availability of large-scaled labeled data. The acquisition of annotated dataset is an expensive and time-consuming task, more so with annotating multi-attributes. To overcome this limitation, we propose a self-supervised learning (SSL) algorithm for multi-modality skin lesion classification. Our algorithm enables the multi-modality learning by maximizing the similarities between paired dermoscopic and clinical images from different views. In addition, we generate surrogate pseudo-multi-labels that represent seven attributes via clustering analysis. We also propose a label-relation-aware module to refine each pseudo-label embedding and capture the interrelationships between pseudo-multi-labels. We validated the effectiveness of our algorithm using well-benchmarked seven-point skin lesion dataset. Our results show that our algorithm achieved better performances than other state-of-the-art SSL counterparts.

\end{abstract}


\section{Introduction}

\begin{figure}[!t]
\centerline{\includegraphics[width=\columnwidth]{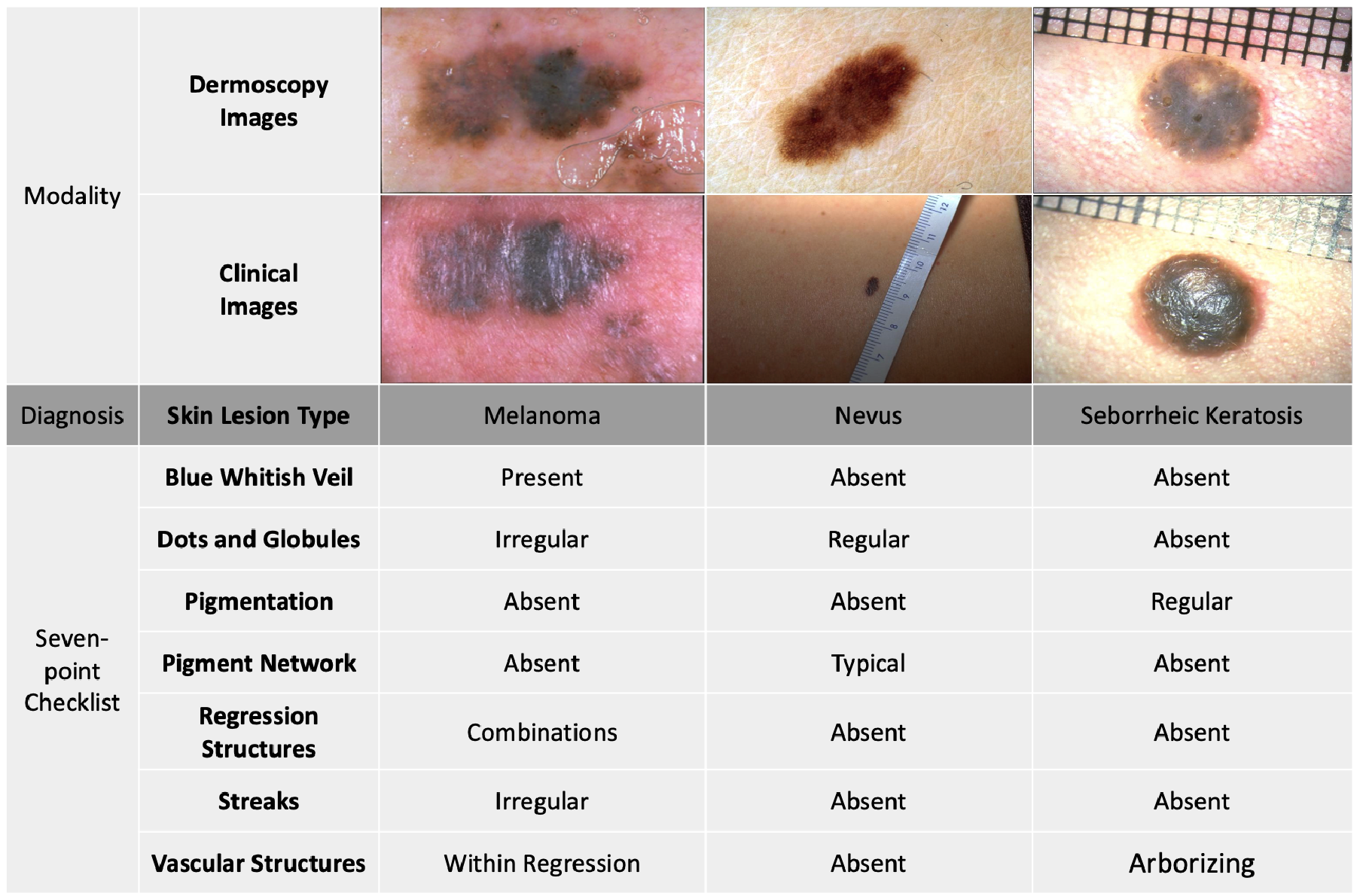}}
\caption{Examples of multi-modal skin lesion images comprising of clinical images and dermoscopy images. Each pair of images is labeled with seven-point checklist and its diagnosis (Melanoma, Nevus or Seborrhic Keratosis).}
\label{fig:skin_example}
\end{figure}

Melanoma is one of the deadliest forms of skin cancer in the world, and the number of incidences has been increasing steadily in recent years \cite{ref1}. Early diagnosis is particularly important as melanoma can be cured with simple excision \cite{ref2}. In a clinical practice, the suspected skin lesions are assessed by examining clinical images and dermoscopy images \cite{ref3}. Clinical images are acquired by a digital camera, showing geometry and color of the skin lesion. On the other hand, dermoscopy images are acquired with a dermatoscope, providing better view of the skin lesion subsurface structures. These combined imaging modalities provide complementary information to assist dermatologists in the diagnosis. Seven-point checklist \cite{ref4} is the most commonly used algorithm for diagnosis, where each attribute in checklist, as denoted in Fig.~\ref{fig:skin_example}, is assigned with 1 point to rate the likelihood of having melanoma. The examined lesion is diagnosed as melanoma when the sum of the score surpasses a given threshold, e.g., greater than three \cite{ref5}. Fig.~\ref{fig:skin_example} gives three example studies showing clinical and dermoscopic images and their corresponding seven attributes. However, such manual examination is time consuming and difficult, particularly for untrained or less experienced dermatologists \cite{ref6}. Moreover, dermatologists are consistently in short supply in rural areas, and consultation costs are rising \cite{ref7}. Motivated by these difficulties, computer-aided diagnosis (CAD) systems have been developed to automate such manual process and provide second opinions to clinicians. In recent years, many CAD systems based on Convolutional Neural Networks (CNNs) have been successful in skin lesion image segmentation and classification \cite{ref8,ref9}. For example, Yu et al. \cite{ref8} proposed to use a very deep CNN with more than 50 layers to acquire richer and more discriminative skin features. However, these methods are limited to using dermoscopy images (a single modality), and therefore discarding useful information contained in the clinical images. More recently, researchers \cite{ref1,ref10} have attempted to develop multi-modality fusion networks to simultaneously learn image features from dermoscopic and clinical images. A common approach in these networks is to use separate CNNs for each imaging modality, with the subsequent step of concatenating the feature outputs from each CNN. Although these studies exploited complementary information from modalities, they directly predicted the diagnosis of melanoma from images without inferring the seven-point checklist, having the likelihood of making a misdiagnosis \cite{ref1}.

Multi-label classification (MLC), where an image can be assigned to multiple classes or labels simultaneously, is a pertinent approach to solve the issue. In MLC setting, the seven attributes are considered as seven labels with the diagnosis as the eighth label for each image. It then learns the interrelationships among the seven attributes and the 8th diagnosis label. For instance, Fu et al. \cite{ref28} proposed a graph-based model to leverage the interrelationships in the seven-point checklist to improve skin lesion classification. However, the performances of these methods are highly dependent on the availability of large-scale labeled training data. Unfortunately, there is a scarcity of large annotated multi-modal skin lesion datasets due to the expensive data acquisition and annotation process. Earlier studies \cite{ref11,ref12} have mitigated this issue by adopting transfer learning such that models, pre-trained with ImageNet \cite{ref13}, can be fine-tuned on the target medical imaging dataset. Despite the effectiveness of transfer learning, there still exists large domain shifts between sources, e.g., ImageNet and skin lesion images \cite{ref14}.

An alternative approach is to use a self-supervised learning (SSL) approach to learn meaningful features using only unlabeled data. Many recent SSL approaches \cite{ref15,ref16,ref17,ref18,ref19,ref20} have been successfully introduced for various tasks in both natural and medical image analysis. For example, Chen et al. \cite{ref15} designed a SSL approach that maximizes similarities between augmented views of the same image and minimizes the similarities with other images. By doing so it provided more robust transfer ability than approaches that used ImageNet-pre-trained weights. Existing SSL-based approaches, however, are not optimized for multi-modality skin lesion images, as they do not consider how image features from different modalities could be fused to complement each other during pre-training process. They are also not able to fully leverage the information embedded within the seven attributes and thus face difficulties in adequately capturing the interrelationships between the attributes \cite{ref1}.
In an attempt to address the issues discussed above, we propose a Self-supervised Multi-Modality learning framework for Multi-label skin lesion classification (SM3). Our contributions are summarized as follows:
\begin{itemize}
    \item We introduce a new SSL pre-training algorithm in which the image features of different modalities are contrasted. Our innovation is to exploit the inherent complementary information within the dermoscopic and clinical images of the same patient which is expected to possess the highest mutual information compared to random pairing between different patients. This is in contrast to existing SSL pre-training methods which are exclusively designed to work with a single imaging modality. Our pre-training task facilitates the fusion of multi-modal image features, thereby fostering improved discrimination and differentiation among various skin lesion classes.
    \item We further innovate in SSL pre-training algorithm designed for MLC. Our approach enables the learning of correlations among the seven attributes and the final diagnosis without using labeled data. To achieve this, we propose a pseudo-multi-labeling scheme (we refer the seven attributes as seven additional labels from now on) that is constructed by multiple cluster analysis for each label, with the centroid of each resulting cluster representing a class label. These pseudo-multi-labels are used to facilitate multi-label self-supervised learning during the pre-training.
    \item Moreover, we improve the learning of correlations by introducing a label-relation-aware module. It uses the distribution of features within pseudo-multi-labels to better capture interrelationships between them. 
    
\end{itemize}

\section{Related Works}
\subsection{Deep Learning based Skin Lesion Classification}
Skin lesion classification has seen significant advancements with the introduction of CNNs \cite{ref21} which has become the preferred technology for developing CAD systems \cite{ref22,ref23,ref24}. In skin lesion classification, CNNs have shown superior performance compared to traditional methods based on handcrafted features \cite{ref11,ref25}. Researchers have extended this approach by using more advanced techniques, such as deep CNNs with more layers \cite{ref8}, attention learning mechanism \cite{ref9}, and regularization strategies for small and unbalanced datasets \cite{ref26}. However, these methods often overlooked clinical images, which are crucial for precise decision making. To address this limitation, Ge et al.\cite{ref10} proposed a multi-modality learning method that utilized both dermoscopy and clinical images by applying separate CNNs for each modality. Subsequent research has built on this approach by incorporating multi-scale feature fusion modules \cite{ref3} and adversarial learning with attention mechanisms \cite{ref27} to capture both correlated and complementary information from two image modalities. In addition, researchers have explored the detection of dermoscopic attributes (multi-label classification) from the seven-point checklist to improve the classification performance \cite{ref1}. For example, Fu et al.\cite{ref28} proposed a graph-based model to capture the interrelationships between different labels. Similarly, Tang et al. \cite{ref29} developed a two-stage learning scheme, where dermoscopy and clinical image features were integrated in the first stage, which were then integrated with patient metadata in the second stage to capture correlations between labels. 

\subsection{SSL in Medical Imaging}
SSL has emerged as a promising alternative to alleviate the problem of expensive and time-consuming annotation processes \cite{ref30}. This is particularly relevant in the context of medical imaging, where annotated data are scarce due to the complicated data acquisition procedures \cite{ref31}. One of common SSL approaches is to use Contrastive Learning \cite{ref51} which uses a pretext task that maximizes similarities between similar data instances while minimizing them with dissimilar instances. For example, it maximizes the similarities between augmented views (e.g., rotated, masked views) of the same image and minimize similarities with augmented views of different images \cite{ref15,ref16,ref17,ref18}. For example, Azizi et al. \cite{ref31} trained a model on an unlabeled dataset and used a pretext task based on multiple images of the same clinical case to improve skin lesion classification. Recently, there was a systematic review that evaluated the use of various SSL algorithms for skin lesion classification \cite{ref32}. There were also other studies of using SSL to address common skin lesion classification challenges, e.g., long-tail out-of-distribution problem \cite{ref33} and light weight models \cite{ref34}. However, all these SSL-based methods focused on using a single medical imaging modality and cannot be directly applied to multi-modality learning. While multi-modality learning can be implemented through the simple concatenation of multi-modality image features, this is not optimal to derive complementary information in a SSL setting. As also pointed by Li et al. \cite{ref35}, naïve concatenation is not an efficient way and could heavily decrease the model performance due to the domain differences between different imaging modalities. To address this issue, Zhang et al. \cite{ref36} proposed to use the Mean Squared Error (MSE) to align multi-modal feature maps and designed a new contrastive loss to enforce the network to focus on the similarities of segmentation masks from paired modalities as well as dissimilarities of unpaired multi-modal data. Huang et al. \cite{ref37} proposed a SSL algorithm for four-modality ultrasound learning, where Mean Absolute Error across different modalities was minimized to ensure that high-level image features extracted from different modalities can be similar. 

\subsection{MLC in Medical Imaging}
MLC considers a scenario where a set of class labels is assigned to a single data instance. In this context, prior research has primarily concentrated on devising models for understanding relationships between labels \cite{ref45}, including approaches such as one-vs-all classifiers \cite{ref47}, tree structures \cite{ref46}, and graph structures \cite{ref48}. Medical imaging field also has adopted MLC, as exemplified by Guan et al.'s use of a residual attention learning framework for chest X-ray image classification. It assigned different weights to different spatial regions based on multi-labels \cite{ref49}. Similarly, Liu et al. \cite{ref50} enhanced model robustness by maintaining the consistency of relationships among different samples under perturbations. Zhang et al. \cite{ref38} employed a triplet attention network with a Transformer to make use of multi-labels together with spatial and category attention features. 

\section{Materials and Method}
\subsection{Materials}
We used a publicly available multi-modality and multi-label skin lesion dataset (Derm7pt) \cite{ref1} in our experiments. It contains a total of 1,011 studies. The dataset is divided into 413 studies for training, 203 studies for validation, and 395 studies for testing, according to \cite{ref1}. Each study contains a pair of dermoscopy and clinical images, a diagnosis (DIAG) label, and seven-point checklist labels. The DIAG label consists of 5 types of skin lesions including Basal Cell Carcinoma (BCC), Nevus (NEV), Melanoma (MEL), Miscellaneous (MISC), and Seborrheic Keratosis (SK). The seven-point checklist labels contain Pigment Network (PN), Blue Whitish Veil (BWV), Vascular Structures (VS), Pigmentation (PIG), Streaks (STR), Dots and Globules (DaG), and Regression Structures (RS). Each seven-point checklist label has different number of classes including Absent (ABS), Typical (TYP), Atypical (ATP), Present (PRS), Regular (REG), and Irregular (IR). The size of dermoscopy images varies from 474 × 512 to 532 × 768 pixels and the clinical images vary from 480 × 512 to 532 × 768 pixels.

\subsection{Preliminaries: SSL Pre-training Strategy}

\begin{figure}[!t]
\centerline{\includegraphics[width=.95\columnwidth]{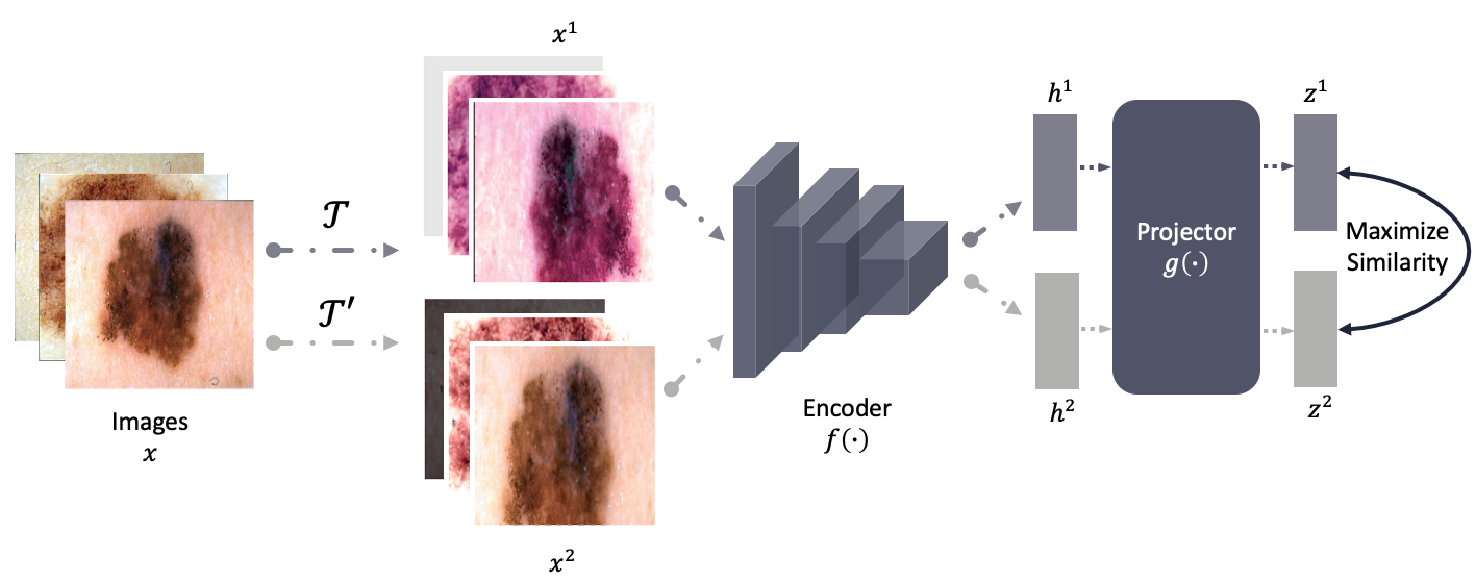}}
\caption{Pipeline of SimCLR applied to skin lesion classification.}
\label{fig:skin_simclr}
\end{figure}

\begin{figure*}[!h]
\centerline{\includegraphics[width=\textwidth]{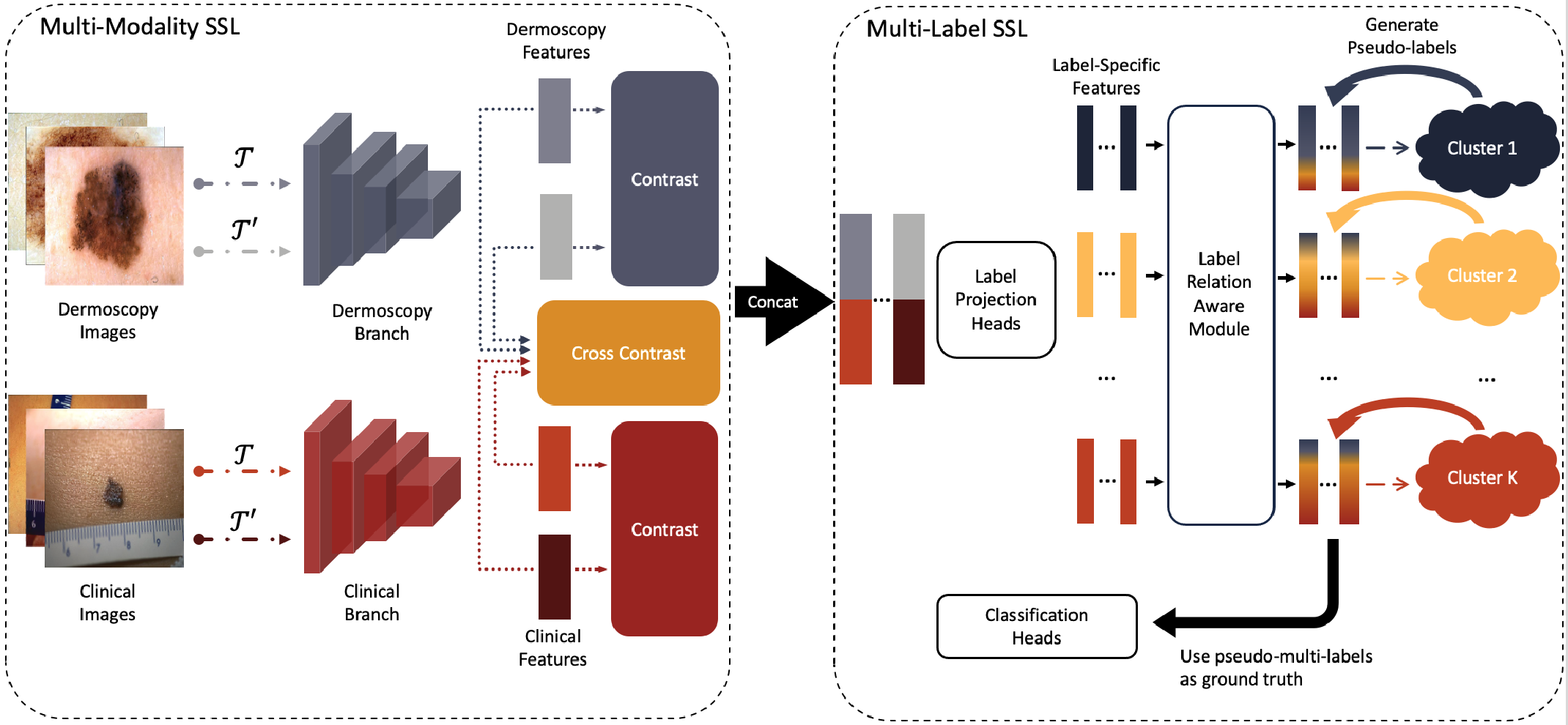}}
\caption{A schematic of SM3. The multi-modality SSL component utilizes two separate branches to extract modality-specific features using SimCLR. A multi-modal fusion is then enabled by contrasting paired dermoscopic and clinical images. In the multi-label SSL component, the concatenated image features are projected into label-specific features, and a label-relation-aware module is applied to learn label correlations in a self-supervised manner. Each label-specific feature is then grouped into similar features and used to generate pseudo-multi-labels. These are used to update the classification heads.}
\label{fig:model_overview}
\end{figure*}

In this work, we used a SSL method SimCLR \cite{ref15} as the base pre-training strategy. The workflow of SimCLR is shown in Fig.~\ref{fig:skin_simclr}. Firstly, two separate data augmentation sets $\{\mathcal{T}, \mathcal{T}'\}$ from the same family of augmentations (including random resized cropping, color jitter, random horizontal flip, and random Gaussian blur) randomly transform any given image sample $x$ into two augmented views $x^1$ and $x^2$, which are considered as a \textit{positive pair}. Then, an encoder network  $f(\cdot)$ extracts image features  $h^1, h^2$ from augmented views, respectively. The choice of encoder network is flexible and can be any CNN architecture. Afterward, a projection head  $g(\cdot)$ maps learned image representations into a latent space $z$. We used the Multi-Layer Perceptron (MLP) as our projection head. Lastly, the contrastive loss is applied to the latent space $z$, aiming to maximize the similarities between positive pairs. The loss function for image $x_i$ is defined as: 
\begin{equation}\label{eq0}
\begin{aligned}
    & \mathcal{L}_{NT-Xent, i} = \\
    & -\log \frac{e{\left(\sigma\left(z_i^1,z_i^2\right)/\tau\right)}}{\sum_{j=1}^{N}{e{\left(\sigma\left(z_i^1,z_j^2\right)/\tau\right)\ }+1_{\left[i \neq j\right]}e{\left(\sigma\left(z_i^1,z_j^1\right)/\tau\right)}}}
\end{aligned}
\end{equation}
where $N$ denotes the batch size, $\tau$ is a temperature hyperparameter. $1_{\left[i \neq j\right]}\in\left\{0,\ 1\right\}$ is an indicator function equaling to 1 if and only if $i \neq j$. $e\left( \cdot \right)$ is the exponentiation operation, and $\sigma \left(u,v\right)=u^\top v/||u||\,||v||$ denotes the cosine similarity function.

\subsection{Overview}

The overview of our method is shown in Fig.~\ref{fig:model_overview}. Initially, modality-specific features from dermoscopic and clinical images were extracted using SimCLR. Subsequently, we pre-trained the multi-modality models by maximizing similarities between paired multi-modality images of the same patient. Following this, the extracted image features were projected into distinct label-specific embedding spaces. A label-relation-aware module was then used to learn correlation between labels. Lastly, we channeled the outputs into clusters, generating pseudo-multi-labels for SSL multi-label pre-training.

\subsection{Self-Supervised Multi-Modality Learning}
Given pairs of dermoscopy and clinical images, we utilized separate CNNs, named dermoscopy branch and clinical branch, to extract corresponding image modality features. These two branches have identical architecture but independent weight updates, which helps to optimize each branch for different image modalities. To enable efficient multi-modality representation learning, three pretext tasks are defined. The first and second pretext task are to apply SimCLR in each model branch to extract specific features from corresponding modalities. We defined the loss functions using Equation~\ref{eq0} as follows:
\begin{equation}\label{eq1}
    \mathcal{L}_{derm}=\mathcal{L}_{NT-Xent}
\end{equation}
\begin{equation}\label{eq2}
    \mathcal{L}_{clinic}=\mathcal{L}_{NT-Xent}
\end{equation}
where $z$ in $\mathcal{L}_{derm}$ comes from dermoscopy images, whereas $z$ in $\mathcal{L}_{clinic}$ is derived from clinical images. We used these loss functions to solve the first and second pretext tasks. In addition, we propose a third task that jointly utilizes both dermoscopy and clinical images, allowing complementary representation learning of the two modalities for multi-modal fusion. The intuition behind our design is that pairs of multi-modality images have more similarities than others, i.e., dermoscopy and clinical images of the same case have maximum mutual information under different augmented views. We implemented this idea by 1) introducing two extra projection heads to map extracted dermoscopy and clinical features into a shared embedding space and 2) maximizing the agreement between randomly augmented views of the same case but different modality data sample by the modified contrastive loss:
\begin{equation}\label{eq3}
\begin{aligned}
    & \mathcal{L}_{mm}=\ -log{\frac{e{\left(\sigma\left(z_i^1,{z^\prime}_i^1\right)/\tau\right)}}{\sum_{j=1}^{N}{e{\left(\sigma\left(z_i^1,{z^\prime}_j^1\right)/\tau\right)\ }+1_{\left[i\neq j\right]}e{\left(\sigma\left(z_i^1,z_j^1\right)/\tau\right)}}}} \\
    & -log{\frac{e{\left(\sigma\left(z_i^1,{z^\prime}_i^2\right)/\tau\right)}}{\sum_{j=1}^{N}{e{\left(\sigma\left(z_i^1,{z^\prime}_j^2\right)/\tau\right)\ }+1_{\left[i\neq j\right]}e{\left(\sigma\left(z_i^1,z_j^1\right)/\tau\right)}}}}
\end{aligned}
\end{equation}
where $z$ is computed from dermoscopy images while $z^\prime$ is calculated from clinical images. We applied these three tasks to train the model by adopting multi-task learning and defined the final loss function as:
\begin{equation}\label{eq4}
    \mathcal{L}_{ssl}=\mathcal{L}_{derm}+\mathcal{L}_{clinic}+\mathcal{L}_{mm}
\end{equation}

\subsection{Self-Supervised Multi-Label Learning}


\textbf{Naïve solution.} Since multiple label predictions are derived from the single image representation and the number of classes is different for each label, we adopted separate classifiers for every label prediction. The classifier $h\left(\cdot\right)$ was built by a label projection head $p\left(\cdot\right)$ and a classification head $q\left(\cdot\right)$ such that $h\left(\cdot\right)=q\left(p\left(\cdot\right)\right)$. Here, $p\left(\cdot\right)$ consists of an MLP aiming to filter label-specific features and $q\left(\cdot\right)$ is a single fully-connected (FC) layer to make final predictions. To enable self-supervised learning of the multi-label classifier, we used the k-means clustering algorithm \cite{ref39} to generate pseudo-multi-labels. We iterated the clustering process independently for $K$ times to produce multiple labels for the same data point where $K$ equals to the number of unique labels in the dataset. We then utilized these generated pseudo-multi-labels to update the parameters of the classifier by optimizing the cross-entropy loss function which is defined as follows:
\begin{equation}\label{eq5}
    \mathcal{L}_{ce}\left(x_i,y_i\right)=\sum_{k=1}^{K}CrossEntropy\left(h_k\left(x_i\right),y_{i,k}\right)
\end{equation}
where $x_i$ is the $i_{th}$ image and $y_i$ is the corresponding pseudo-multi-labels containing $\{y_{i,1},\ldots,y_{i,K}\}$. $h_k\left(\cdot\right)$ denotes the $k_{th}$ classifier.

\textbf{Label-relation-aware solution.} The above naïve solution, however, yielded degraded results in our preliminary experiments (Section V.B.2), which overlooked the relationships between labels. We therefore further refined each label embedding by understanding the relationships between other label embeddings guided by an attention mechanism \cite{ref40}. Attention mechanism has been successfully used to capture relationships between feature representations. Formally, we inserted a function $W\left(\cdot\right)$ before $q\left(\cdot\right)$ and fed outputs of all $p\left(\cdot\right)$ into it. Then, we rewrote the classifier function as:
\begin{equation}\label{eq6}
    h\left(\cdot\right)=q\left(W\left(p_1\left(\cdot\right),\ldots,p_K\left(\cdot\right)\right)\right)
\end{equation}
where $\{p_1,\ldots,p_K\}$ are label projection heads for $K$ unique labels. Here, the features specific to each label only contain information about that label. However, our label-relation-aware module come into play to connect all labels information and learn the relationships between them. As a result, the prediction of a single label involves contributions from other label information. In this work, we used $W\left(\cdot\right)$ as the encoder layer of Transformer \cite{ref40}.

\subsection{Inference Pipeline}

\begin{figure}[!t]
\centerline{\includegraphics[width=\columnwidth]{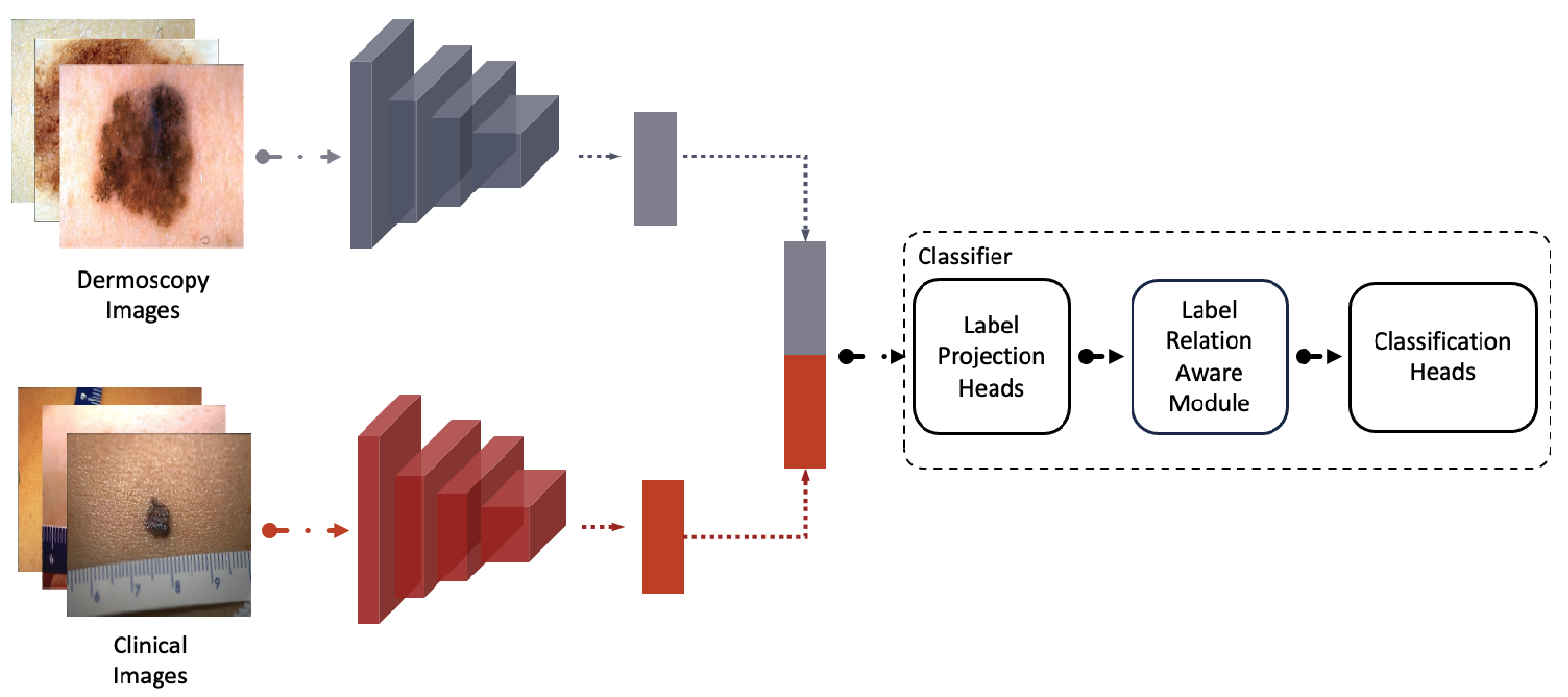}}
\caption{Pipeline of the inference module. Initially, it employs two distinct branches to extract features from dermoscopic and clinical images separately. These extracted features are subsequently fused by concatenation. The combined feature set is then passed to a subsequent classifier for generating the final prediction.}
\label{fig:inference_pipeline}
\end{figure}

We adopted a vanilla implementation of multi-modality and multi-label classification scheme (as shown in Fig.~\ref{fig:inference_pipeline}) to emphasize the effectiveness of our proposed SSL pre-training method. Following the design of multi-modality model fusion in preliminary work EmbeddingNet \cite{ref41}, we applied two separate branches to extract dermoscopic and clinical features. These two modality features were then concatenated and fed into a subsequent classifier to make the final predictions. This model was referred as the Baseline in the Section IV. Experiments, and we initialized both the branches and the classifier using our SM3 pre-trained weights.

\section{Experiments}

\subsection{Experiment Configurations}
We used a deep learning library PyTorch \cite{ref42} to implement our algorithm. All the experiments were conducted on two NVIDIA RTX 3080Ti 12GB GPUs. For fair comparisons, we used ResNet-50 \cite{ref24} as the CNN backbone since ResNet-50 has been commonly applied for various skin lesion classification and segmentation tasks including \cite{ref3,ref27,ref28,ref29}. The output dimension of the projectors in contrastive learning was set to 128. For multi-label SSL, we set the label projection head as a single-layer MLP with a dimension of 512. We used a single Transformer encoder layer with a head of 1, a feed forward dimension of 128 and a dropout rate of 0.1 for our label-relation-aware module . All these hyperparameters were chosen based on our empirical studies. We resized all of images into 224 × 224. Code is available at \url{https://github.com/Dylan-H-Wang/skin-sm3}. 

\subsubsection{Pre-training}
For multi-modality SSL, we set the batch size to 96, learning rate to 1e-6, the number of epoch to 400, and temperature $\tau$ to 0.1. For multi-label SSL, we used batch size of 256, learning rate of 1e-4 and the number of epoch as 150. AdamW \cite{ref43} was used as the optimizer with default parameters ($\beta_1=0.9$, $\beta_2=0.999$, and weight decay $= 0.01$). We followed the commonly used SimCLR data augmentation design during the pre-training.

\begin{table*}[t!]
\centering
\caption{Classification performance evaluated by AUC, Sensitivity (Sens), Specificity (Spec), and Precision (Prec) on the Derm7pt dataset. Three settings are presented: Supervised, SSL + Linear Probing, and SSL + Fine-tuning. The best performances in the four metrics are shown with different superscripts (AUC-$\ast$, Sens-$\vartriangle$, Spec-$\lozenge$, and Prec-$\square$). The best results in the supervised setting are italic, the best results in the linear probing setting are underlined, and the best results in fine-tuning are in bold.}
\label{tab:main_exp}
\resizebox{.9\textwidth}{!}{%
\begin{tabular}{@{}llllllllllll@{}}
\toprule
 &  &  & BWV & DaG & PIG & PN & RS & STR & VS & DIAG &  \\ \cmidrule(lr){4-11}
\multirow{-2}{*}{Strategy} & \multirow{-2}{*}{Method} & \multirow{-2}{*}{Metrics} & PRS & IR & IR & ATP & PRS & IR & IR & MEL & \multirow{-2}{*}{AVG} \\ \midrule
 &  & AUC & 89.2 & 79.9 & 79.0 & 79.9 & 82.9 & 78.9 & 76.1 & 86.3 & 81.5 \\
 &  & Sens. & 77.3 & 62.1 & 59.7 & 48.4 & 66.0 & 51.1 & 13.3 & 61.4 & 54.9 \\
 &  & Spec. & 89.4 & 78.9 & 80.1 & 90.7 & 81.3 & 85.7 & 97.5 & 88.8 & 86.6 \\
 & \multirow{-4}{*}{Inception-Combined} & Prec. & 63.0 & 70.5 & 57.8 & 61.6 & 56.5 & 52.7 & 30.8 & 65.3 & 57.3 \\ \cmidrule(l){3-12} 
 &  & AUC & 89.8 & 82.6 & 81.3 & 78.3 & 81.9 & 77.6 & 82.7 & 85.6 & 82.5 \\
 &  & Sens. & 92.2 & 80.2 & 55.7 & 40.9 & 95.2 & 35.1 & 20.0 & 68.8 & \textit{61.0$^\vartriangle$} \\
 &  & Spec. & 65.3 & 71.6 & 86.3 & 92.4 & 41.5 & 90.0 & 98.4 & 85.4 & 78.9 \\
 & \multirow{-4}{*}{HcCNN} & Prec. & 91.9 & 69.6 & 65.1 & 62.3 & 81.6 & 52.4 & 50.0 & 54.5 & 65.9 \\ \cmidrule(l){3-12} 
 &  & AUC & 90.8 & 83.1 & 83.6 & 87.5 & 79.0 & 81.2 & 75.4 & 87.6 & 83.5 \\
 &  & Sens. & 69.9 & 70.1 & 39.2 & 77.5 & 21.9 & 67.0 & 3.6 & 59.0 & 51.0 \\
 &  & Spec. & 91.0 & 78.8 & 95.8 & 79.0 & 96.8 & 80.3 & 100.0 & 89.5 & 88.9 \\
 & \multirow{-4}{*}{GIIN} & Prec. & 67.4 & 74.9 & 82.3 & 48.4 & 73.5 & 50.4 & 100.0 & 65.6 & \textit{70.3$^\square$} \\ \cmidrule(l){3-12} 
 &  & AUC & 91.1 & 81.9 & 83.4 & 82.0 & 86.7 & 80.7 & 80.9 & 89.1 & \textit{84.5$^\ast$} \\
 &  & Sens. & 75.0 & 66.7 & 67.9 & 58.5 & 72.1 & 57.3 & 0.0 & 65.8 & 57.9 \\
 &  & Spec. & 90.3 & 82.4 & 83.0 & 85.6 & 82.6 & 85.9 & 92.4 & 91.4 & 86.7 \\
 & \multirow{-4}{*}{AMFAM} & Prec. & 56.0 & 82.5 & 61.3 & 51.6 & 46.2 & 54.3 & 0.0 & 76.2 & 53.5 \\ \cmidrule(l){3-12} 
 &  & AUC & 90.6 & 80.1 & 83.5 & 83.9 & 81.7 & 81.4 & 78.9 & 89.0 & 83.6 \\
 &  & Sens. & 66.7 & 68.4 & 58.9 & 49.5 & 47.1 & 47.9 & 20.0 & 62.4 & 52.6 \\
 &  & Spec. & 91.6 & 72.9 & 87.1 & 90.1 & 96.2 & 88.4 & 97.8 & 88.8 & 89.1 \\
 & \multirow{-4}{*}{F4M-FS} & Prec. & 64.9 & 67.2 & 67.6 & 60.5 & 82.0 & 56.2 & 42.9 & 65.6 & 63.4 \\ \cmidrule(l){3-12} 
 &  & AUC & 87.4 & 78.0 & 82.0 & 79.4 & 80.6 & 76.1 & 80.5 & 86.7 & 81.3 \\
 &  & Sens. & 49.3 & 66.1 & 46.8 & 40.9 & 34.0 & 47.9 & 3.3 & 51.5 & 42.5 \\
 &  & Spec. & 98.4 & 72.0 & 87.8 & 86.1 & 94.5 & 93.4 & 99.7 & 95.6 & \textit{90.9}$^\lozenge$ \\
\multirow{-24}{*}{Supervised} & \multirow{-4}{*}{Baseline-50-ImageNet} & Prec. & 88.1 & 65.7 & 63.7 & 47.5 & 69.2 & 69.2 & 50.0 & 80.0 & 66.7 \\ \midrule
 &  & AUC & 85.6 & 76.1 & 77.0 & 75.4 & 78.7 & 72.8 & 76.2 & 83.6 & 78.2 \\
 &  & Sens. & 20.0 & 70.1 & 46.0 & 26.9 & 12.3 & 24.5 & 0.0 & 49.5 & 31.1 \\
 &  & Spec. & 99.7 & 69.7 & 85.2 & 94.4 & 97.2 & 94.0 & 100.0 & 91.8 & 91.5 \\
 & \multirow{-4}{*}{SimCLR} & Prec. & 93.8 & 65.3 & 58.8 & 59.5 & 61.9 & 56.1 & 0.0 & 67.6 & 57.9 \\ \cmidrule(l){3-12} 
 &  & AUC & \multicolumn{1}{c}{86.7} & \multicolumn{1}{c}{76.8} & \multicolumn{1}{c}{81.5} & \multicolumn{1}{c}{79.1} & \multicolumn{1}{c}{77.1} & \multicolumn{1}{c}{79.2} & \multicolumn{1}{c}{69.7} & \multicolumn{1}{c}{86.6} & \multicolumn{1}{c}{79.6} \\
 &  & Sens. & \multicolumn{1}{c}{0.0} & \multicolumn{1}{c}{66.7} & \multicolumn{1}{c}{9.7} & \multicolumn{1}{c}{0.0} & \multicolumn{1}{c}{0.0} & \multicolumn{1}{c}{0.0} & \multicolumn{1}{c}{0.0} & \multicolumn{1}{c}{2.0} & \multicolumn{1}{c}{9.8} \\
 &  & Spec. & \multicolumn{1}{c}{100.0} & \multicolumn{1}{c}{74.3} & \multicolumn{1}{c}{99.3} & \multicolumn{1}{c}{99.7} & \multicolumn{1}{c}{100.0} & \multicolumn{1}{c}{100.0} & \multicolumn{1}{c}{100.0} & \multicolumn{1}{c}{100.0} & \multicolumn{1}{c}{\begin{tabular}[c]{@{}c@{}}{\ul 96.7$^\lozenge$}\end{tabular}} \\
 & \multirow{-4}{*}{SSD-KD} & Prec. & \multicolumn{1}{c}{0.0} & \multicolumn{1}{c}{67.8} & \multicolumn{1}{c}{85.7} & \multicolumn{1}{c}{0.0} & \multicolumn{1}{c}{0.0} & \multicolumn{1}{c}{0.0} & \multicolumn{1}{c}{0.0} & \multicolumn{1}{c}{100.0} & \multicolumn{1}{c}{31.7} \\ \cmidrule(l){3-12} 
 &  & AUC & 90.4 & 76.5 & 80.4 & 78.2 & 77.4 & 75.4 & 79.4 & 85.0 & {\ul 80.4$^\ast$} \\
 &  & Sens. & 50.7 & 40.1 & 31.5 & 39.8 & 18.9 & 41.5 & 0.0 & 56.4 & \begin{tabular}[c]{@{}l@{}}{\ul 34.9$^\vartriangle$}\end{tabular} \\
 &  & Spec. & 95.9 & 89.9 & 94.5 & 89.4 & 99.3 & 86.0 & 100.0 & 93.9 & {\color[HTML]{000000} 93.6} \\
\multirow{-12}{*}{SSL + Linear Probing} & \multirow{-4}{*}{SM3-linear} & Prec. & 74.5 & 76.3 & 72.2 & 53.6 & 90.9 & 48.1 & 0.0 & 76.0 & {\ul 61.5$^\square$} \\ \midrule
 &  & AUC & 91.1 & 81.9 & 82.8 & 78.0 & 82.5 & 79.8 & 81.2 & 86.1 & 82.9 \\
 &  & Sens. & 34.7 & 51.4 & 68.5 & 32.3 & 25.5 & 48.9 & 20.0 & 50.5 & \textbf{\begin{tabular}[c]{@{}l@{}}41.5$^\vartriangle$\end{tabular}} \\
 &  & Spec. & 99.1 & 90.4 & 80.4 & 94.7 & 96.9 & 89.7 & 96.4 & 93.5 & 92.6 \\
 & \multirow{-4}{*}{SM3-finetune} & Prec. & 89.7 & 81.3 & 61.6 & 65.2 & 75.0 & 59.7 & 31.6 & 72.9 & 67.1 \\ \cmidrule(l){3-12} 
 &  & AUC & 92.9 & 80.3 & 84.5 & 82.3 & 84.2 & 84.5 & 77.8 & 90.1 & \textbf{84.6$^\ast$} \\
 &  & Sens. & 70.7 & 52.5 & 48.4 & 35.5 & 45.3 & 37.2 & 3.3 & 31.7 & 40.6 \\
 &  & Spec. & 92.5 & 85.3 & 91.9 & 92.4 & 94.5 & 91.9 & 100.0 & 98.0 & \textbf{\begin{tabular}[c]{@{}l@{}}93.3$^\lozenge$\end{tabular}} \\
\multirow{-8}{*}{SSL + Fine-tuning} & \multirow{-4}{*}{FM4-FS + SM3} & Prec. & 68.8 & 74.4 & 73.2 & 58.9 & 75.0 & 73.2 & 100.0 & 84.2 & \textbf{76.0$^\square$} \\ \bottomrule
\end{tabular}%
}
\end{table*}

\subsubsection{Linear Probing and Fine-tuning}
We adopted linear probing protocol where CNNs were frozen and only the classifier was fine-tuned \cite{ref16}. We set learning rate as 1e-3, batch size as 128 and the number of epochs as 50. The optimizer was AdamW with default parameters as in the pre-training. We used data augmentations including random resized crop and random horizontal flip. We also evaluated the non-linear quality of learned representations \cite{ref44} such that CNNs were initialized with pre-trained weights and fine-tuned with all layers. We set learning rate as 1e-4, batch size as 64 and the number of epochs as 50. The settings of optimizer and data augmentation were the same as linear probing experiments.

\subsection{Evaluation Setups}
\subsubsection{Performance metrics}
The model performances were evaluated using metrics including area under receiver operating characteristic curve (AUC), sensitivity (Sens), speciﬁcity (Spec), and precision (Prec).

\subsubsection{Comparison to the state-of-the-arts}
The Baseline model is a vanilla implementation of multi-modality and multi-label classification as shown in Fig.\ref{fig:inference_pipeline}. We setup the upper bound of linear probing experiments by initializing the Baseline with ImageNet-pre-trained weights (Baseline-50-ImageNet). We also initialized the Baseline using our proposed SM3 pre-trained weights and conducted linear probing (SM3-linear) and fine-tuning (SM3-fine-tune) experiments. We benchmarked state-of-the-art (SOTA) SSL methods, including general SSL method SimCLR \cite{ref15} and SSL method optimized for dermoscopic image analysis named SSD-KD \cite{ref34}, along with supervised SOTAs including commonly used baseline Inception-combined \cite{ref1} and HcCNN \cite{ref3}, and recent graph-based GIIN \cite{ref28}, adversarial-based AMFAM \cite{ref27} and patient-meta-based FM4-FS \cite{ref29}. In addition, we initialized FM4-FS with our SM3 (FM4-FS + SM3) and fine-tuned with all layers.

\subsubsection{Ablation studies}
For each of the ablation studies, hyperparameters were grid-searched and the metrics were based on the AUC scores. We conducted ablation studies on the Multi-Modality SSL (MMSSL) and the Multi-Label SSL (MLSSL) components by incorporating them into the Baseline separately. For MMSSL, we evaluated three different fusion strategies including: i) \textit{concat}: concatenating dermoscopic and clinical features and maximizing the similarities between different views of the concatenation; ii) \textit{sep\_shared}:  maximizing the similarities between views of paired dermoscopy and clinical images using a shared projection head $g\left(\cdot \right)$; iii) \textit{sep\_sep}: same as \textit{sep\_shared} but applying separate projection heads for each modality. We also used weights pre-trained on ImageNet (\textit{ImageNet}) and weights pre-trained by SimCLR (\textit{SimCLR}) as comparisons. The weights pre-trained on ImageNet were downloaded from Torchvision\footnote{https://github.com/pytorch/vision} and initialized to each branch. SimCLR-pre-trained-weights were obtained by using SimCLR pre-train each branch on the corresponding image modality separately.

For MLSSL, we evaluated five different strategies: i) \textit{no\_proj}: there was no label projection head; ii) \textit{proj} : naively applying a label projection head for each label; iii) \textit{msa}: applying the multi-head self-attention; iv) \textit{tel}: applying a Transformer encoder layer; v) \textit{te}: applying a Transformer based encoder.

In addition, we conducted pair match experiments that query clinical images using dermoscopic images as the keys. It aims to find matching corresponding clinical images. It helps to explore how different MMSSL fusion strategies utilize the mutual information between two modalities. Based on the assumption that paired dermoscopic and clinical image features contain the highest similarity, we generated a cosine similarity score matrix for ranking. We assessed the top-1 accuracy (Acc@1) that determines whether paired image features have the highest similarity score, and the top-5 accuracy (Acc@5) that considers the top five highest scores. Moreover, the average rank was computed by averaging the ranks of each paired image's similarity score relative to others. By calculating these metrics, we quantify the extent to which complementary information between two modalities were effectively leveraged by different strategies, and thus guiding the selection of an optimized method for efficient multi-modality fusion.

\section{Results}

\subsection{Comparisons to The State-of-the-arts}
The primary results of the experiment are presented in Table~\ref{tab:main_exp}. We first evaluated the effectiveness of the proposed SM3 representations via linear probing (SSL + Linear Probing). When compared to SimCLR, our SM3 showed consistent improvements including a 2.2\% increase in mean AUC, 3.8\% increase in mean Sens, 2.1\% increase in mean Spec, and 3.6\% in mean Prec, although SimCLR performed better in some categories, e.g., RS category. SSD-KD, based on SSL and knowledge distillation, achieved a second-best performance with an AUC of 79.6, mean Sens of 9.8, mean Spec of 96.7, and mean Prec of 31.7. In comparison, our SM3 had much higher mean Sens (+25.1\%), mean Prec (+29.8\%) and relatively higher mean AUC (+0.8\%) but lower mean Sens (-3.1\%). To understand the upper bound of our SSL linear probing, we compared the SM3-initialized Baseline model with the supervised ImageNet-initialized Baseline model, with SM3 resulting in a higher mean Specificity (+2.7\%), a small gap in terms of mean AUC (-0.9\%), mean Sensitivity (-7.6\%), and mean Precision (-5.2\%). 

We conducted experiments to assess the effectiveness of SM3 in improving various backbones (SSL + Fine-tuning). After fine-tuning the SM3-initialized Baseline model, the mean AUC improved from 81.3 to 82.9, surpassing both Inception-Combined (mean AUC of 81.5) and HcCNN (mean AUC of 82.5). This improvement was consistent regarding mean Spec and mean Prec. We also experimented with another existing method, F4M-FS by replacing the ImageNet-pre-train weights with our SM3-pre-trained weights while keeping other components unchanged. We found that the SM3-initialized FM4-FS achieved 1\% increase in mean AUC, 4.2\% in mean Spec and 12.6\% in mean Prec. Compared to the current supervised state-of-the-art AMFAM which obtained a mean AUC of 84.5, mean Sens of 57.9, mean Spec of 86.7, and mean Prec of 53.5, SM3-initialized FM4-FS outperformed it by 0.1\% in mean AUC, 6.6\% in mean Spec and 22.5\% in mean Prec but with 17.3\% lower in mean Sens.

\subsection{Ablation studies}

\begin{table}[]
\centering
\caption{Ablation on multi-modality and multi-label SSL. The metrics are based on mean AUC scores and the best results are in bold.}
\label{tab:ab_mmssl_mlssl}
\resizebox{\columnwidth}{!}{%
\begin{tabular}{@{}llllllllll@{}}
\toprule
\multirow{3}{*}{Strategy} & \multicolumn{9}{c}{AUC} \\ \cmidrule(l){2-10} 
 & BWV & DaG & PIG & PN & RS & STR & VS & DAIG & \multirow{2}{*}{AVG} \\ \cmidrule(lr){2-9}
 & PRS & IR & IR & ATP & PRS & IR & IR & MEL &  \\ \midrule
ImageNet & 83.4 & 72.0 & 77.6 & 78.0 & 73.0 & 73.0 & 79.6 & 82.7 & 77.4 \\
SimCLR & 85.6 & 76.1 & 77.0 & 75.4 & 78.7 & 72.8 & 76.2 & 83.6 & 78.2 \\
MMSSL$_{concat}$ & 86.2 & 76.0 & 74.2 & 74.5 & 76.5 & 70.9 & 74.0 & 83.6 & 77.0 \\
MMSSL$_{sep\_shared}$ & 88.2 & 75.3 & 75.6 & 75.1 & 74.7 & 71.7 & 75.5 & 83.9 & 77.5 \\
MMSSL$_{sep\_sep}$ & 87.7 & 75.2 & 79.4 & 77.8 & 77.8 & 73.4 & 79.8 & 84.6 & \textbf{79.5} \\ \midrule
\multicolumn{10}{l}{} \\ \midrule
MLSSL$_{no\_proj}$ & 89.5 & 75.3 & 79.6 & 77.2 & 77.8 & 73.1 & 75.2 & 83.5 & 78.9 \\
MLSSL$_{proj}$ & 86.9 & 75.7 & 76.3 & 76.1 & 77.6 & 72.1 & 77.1 & 84.1 & 78.2 \\
MLSSL$_{msa}$ & 89.3 & 75.3 & 79.7 & 74.2 & 78.0 & 73.8 & 78.8 & 85.7 & 79.3 \\
MLSSL$_{tel}$ & 90.4 & 76.5 & 80.4 & 78.2 & 77.4 & 75.4 & 79.4 & 85.0 & \textbf{80.4} \\
MLSSL$_{te}$ & 90.0 & 75.3 & 80.2 & 78.6 & 78.3 & 72.5 & 74.9 & 84.5 & 79.3 \\ \bottomrule
\end{tabular}%
}
\end{table}

\subsubsection{Efficacy of multi-modality SSL}
The ablation results of MMSSL are presented in Table~\ref{tab:ab_mmssl_mlssl}. Compared to the ImageNet-pre-trained weights (mean AUC of 77.4), \textit{SimCLR} strategy achieved a higher score with a mean AUC of 78.2. Compared to the \textit{SimCLR}, \textit{concat} strategy resulted in a decreased mean AUC of 77. The strategy \textit{sep\_shared} helped to improve the performance by 0.5\% compared to naïve concatenation. In contrast, only \textit{sep\_sep} strategy, which maximized the mutual information between paired dermoscopic and clinical images with separate projection heads, resulted in an improved model performance with a mean AUC of 79.5 (increased by 1.3\% compared to \textit{SimCLR}).

\subsubsection{Efficacy of multi-label SSL}
The choice of MLSSL strategies affected performance differently as shown in Table~\ref{tab:ab_mmssl_mlssl}. Firstly, we assessed when there was no label projection head (\textit{no\_proj}) applied, i.e., using concatenated multi-modality features to generate pseudo-multi-labels. We found that without label projection, the pre-trained multi-label classifier could not learn meaningful representations, which decreased the mean AUC from 79.5 to 78.9. Moreover, naively applying a label projection head for each label (\textit{proj}), without the proposed label-relation-aware module, resulted in worse representations due to the ignorance of structure and relationships among labels, and resulting in decreased mean AUC by an additional 0.7\%. The multi-head self-attention (\textit{msa}) was capable of learning and building correlations among labels during SSL pre-training such that it can achieve a similar mean AUC of 79.3. We also evaluated the inclusion of a Transformer encoder layer (\textit{tel}) to measure the benefit of label associations. This strategy improved the result by 0.9\%. The use of different Transformer based encoder (\textit{te}) did not improve the performances.

\subsubsection{Pair matching between different modalities}
The pair matching results of different fusion strategies are shown in Table~\ref{tab:ab_pair_match}. The \textit{sep\_sep} strategy outperformed others by a large margin, with the average rank surpassing that of the \textit{ImageNet} strategy (97.67) by approximately 90 points, achieving top 1\% (7.23/413) rank. The accuracy metrics followed the same trend with \textit{sep\_sep} strategy obtaining the highest Acc@1 of 0.42 and Acc@5 of 0.73. Strategies without multi-modality pre-training generally had worse performance in pair matching, for example, \textit{ImageNet} had average rank of 97.67, Acc@1 of 0.21, and Acc@5 of 0.37, and  \textit{SimCLR} obtained average rank of 83.16, Acc@1 of 0.22 and Acc@5 of 0.38. Additionally, naïve concatenation did not aid in mutual information learning with it achieving average rank of 90.55, Acc@1 of 0.2 and Acc@ 5 of 0.34. Analogously, directly contrasting the two modality images boosted the performance of average rank by 76.86, Acc@ of 0.22, and Acc@5 of 0.23.

\begin{table}[]
\centering
\caption{Ablation on pair matching between different image modalities for different SSL multi-modality strategies. The metrics are based on mean AUC scores and the best results are in bold.}
\label{tab:ab_pair_match}
\resizebox{\columnwidth}{!}{%
\begin{tabular}{@{}lclllclllc@{}}
\toprule
Strategy & \begin{tabular}[c]{@{}c@{}}Avg Rank\\ (total 413)\end{tabular} &  &  &  & Acc@1 &  &  &  & Acc@5 \\ \midrule
ImageNet & 97.67 &  &  &  & 0.21 &  &  &  & 0.37 \\
SimCLR & 83.16 &  &  &  & 0.22 &  &  &  & 0.38 \\
MMSSL$_{concat}$ & 90.55 &  &  &  & 0.20 &  &  &  & 0.34 \\
MMSSL$_{sep\_shared}$ & 13.69 &  &  &  & 0.32 &  &  &  & 0.57 \\
MMSSL$_{sep\_sep}$ & 7.23 &  &  &  & 0.42 &  &  &  & 0.73 \\ \bottomrule
\end{tabular}%
}
\end{table}

\section{Discussion}
The main findings are that: 1) Our SM3 had superior performances compared with other SOTA SSL methods in a multi-modality and multi-label setting; 2) SM3 was shown to be effective in improving other existing methods, outperforming ImageNet-pre-trained counterparts; 3) the ablation studies showed that both the MMSSL and MLSSL components contributed to the overall performance improvements.

\subsection{Comparisons to The State-of-the-arts}
As shown in Table~\ref{tab:main_exp}, most methods performed relatively well due to the use of complex multi-modality fusion techniques, such as class-balanced sampling and multi-task loss in Inception-Combined \cite{ref1}, concatenation of intermediary image features in HcCNN \cite{ref3}, adversarial fusion with attention mechanism in AMFAM \cite{ref27}, and hierarchy fusion at the feature and decision levels in F4M-FS \cite{ref29}. However, most of these methods ignored the MLC setting and did not exploit the interrelationships among labels. GIIN \cite{ref28}, on the other hand, additionally leveraged a graph module to model the label relationships which resulted in improved performance. Compared to Baseline-50-ImageNet, our SM3-fine-tuned method achieved a 1.6\% increase in mean AUC, surpassing established multi-modal fusion strategies like Inception-Combined and HcCNN. Similarly, our SM3 increased the recently published F4M-FS performance with an improved mean AUC of 1\%. This indicates that SM3-pre-trained weights could be an alternative to ImageNet-pre-trained weights for improving the generalizability of different methods in skin lesion analysis.

In linear probing experiments, our method demonstrated a mean AUC enhancement of 2.2\% when compared to the established SimCLR \cite{ref15}. Furthermore, in comparison to the recent SSL skin lesion classification approach of SSD-KD \cite{ref34}, SM3 exhibited a mean AUC increase of 0.8\%. It is worth noting that the degradation in performance with mean Sens metric is expected. This is mainly attributed to the fact that a large proportion of contributions are coming from the VS category, which tends to have low Sens with a relatively high Spec. As indicated by Kawahara et al. \cite{ref1}, the low sensitivity and high specificity is likely attributed to the substantial class imbalance issue in the dataset, where there only exists 71 IR studies in the VS category out of 1,011 studies. By pre-training the model on a skewed dataset without labels, this discrepancy was accumulated resulting in a lower Sens score with a higher Spec score. Nevertheless, we identified that SM3 was effective in dealing with imbalanced data in the seven-point dataset when compared to the other two SSL methods. For instance, IR in the VS category has extremely limited number of cases when compared to the more prevalent REG, and both compared methods struggled to correctly classify IR. In contrast, our method leveraged multi-modality and multi-label techniques managing to correctly classify this minority class. These results underscore the efficacy of our multi-modality SSL design, which harnesses the supplementary potential of dermoscopic and clinical image features to extract enhanced discriminative skin attributes. Moreover, the inclusion of self-supervised multi-label pre-training augments the model's capacity to glean intricate interrelationships among labels, thereby contributing to a more consistent and reliable classification performance.

\subsection{Ablation Studies}
We found that naïve SSL pre-training (i.e., \textit{SimCLR} strategy) with multi-modal data contributed to improving the baseline performance when compared to the commonly used ImageNet-pre-trained weights. This finding is consistent with previous work by Menegola et al \cite{ref14}, where the domain gap between natural images and skin lesion images degraded performance. The \textit{SimCLR} strategy, however, is not optimal for multi-modality learning since the mutual information between two modalities are not leveraged during the pre-training. Nevertheless, for the multi-modal SSL, inappropriate multi-modal fusion (i.e., \textit{concat}) could hinder the learning of meaningful representations. This observation is consistent with findings from a previous work \cite{ref35} that naïve concatenation may result in a worse performance due to domain shift among different modalities. For example, the strategy \textit{sep\_shared} was better than \textit{concat} but still inferior to the \textit{SimCLR} strategy in terms of the mean AUC. In contrast, the \textit{sep\_sep} strategy demonstrated higher performance overall and we attribute this to the application of separate projection heads which can focus on independent image modality features and decide how to map them to increase their similarities. Furthermore, we also identified that contrasting naïve concatenation (\textit{concat}) did not contribute to the learning of mutual information between the two modalities, and directly contrasting two modality features (\textit{sep\_shared} and \textit{sep\_sep}) was more effective giving better pair matching performance. It is noteworthy that although \textit{sep\_shared} was capable of learning more mutual information than \textit{concat} and \textit{SimCLR} strategies, its classification accuracy was lower than that of \textit{SimCLR}. This suggests that an inefficient multi-modality pre-training, i.e., sharing the same projection head, learns trivial complementary multi-modality information and hinders the extraction of individual modality features.

In addition, we observed that directly pre-training a multi-label classifier on image features without label projection heads (\textit{no\_proj}) was not helpful, and simply projecting image features into label embeddings (\textit{proj}) can disrupt self-supervised multi-label learning. We hypothesize that such failure was caused by the independent learning of label projection heads. A naïve label correlation learner (\textit{msa}) had trivial contributions for multi-label SSL, whereas complex model (\textit{te}) cannot achieve reasonable results neither attributing to the fact that the performance of the Transformer based architecture is heavily reliant on the use of large training dataset, which cannot be satisfied with the current experimental dataset. Therefore, it is essential to define an optimal learning strategy, i.e., the Transformer encoder layer, for learning the correlations among labels. 

\subsection{Limitations and Future Works}
Pre-training on an imbalanced dataset, without labels, can accumulate to discrepancies, such that it results in a more severe class imbalance issue. To mitigate this, we will consider developing pre-training models with consideration of imbalanced dataset by combining predictions from multiple models trained on different subsets of the dataset and focusing on minority class samples. Although our method was demonstrated for skin lesion dataset, we suggest that it is applicable to other multi-modality and multi-label imaging datasets where it retains informative mutual features between modalities and relationships between labels, such as with classification on multi-modality PET-CT and PET-MR datasets which contains two imaging modalities with complimentary information.

\section{Conclusion}
In this paper, we introduced a new SSL algorithm for multi-modality and multi-label skin lesion classification. Specifically, maximum complementary information between dermoscopic and clinical images were leveraged during the pre-training when we directly contrasted these two modalities with separate projection heads. Experiments showed that this multi-modality SSL scheme can improve the accuracy of skin lesion classification. Furthermore, we found that generating pseudo-multi-labels using clustering analysis was a surrogate solution for self-supervised multi-label training. With our label-relation-aware module, SM3 was able to capture the interrelationships between labels. Our SM3 outperformed other SOTA SSL methods and helped to improve existing methods by using SM3-pre-trained weights.


\bibliographystyle{IEEEtran}
\bibliography{ref.bib}
\end{document}